\documentclass{svproc}

\usepackage{url}

\usepackage{xcolor}
\usepackage{soul}

\usepackage{todonotes}
\presetkeys{todonotes}{inline,backgroundcolor=pink,caption={}}{}

\usepackage[colorlinks=true,allcolors=red]{hyperref}

\usepackage{amsmath}
\usepackage{amssymb}

\begin{document}

\mainmatter 

\title{Semantic-based Loco-Manipulation for Human-Robot Collaboration in Industrial Environments}

\titlerunning{Semantic-based Loco-Manipulation}  

\author{Federico Rollo\inst{1,2,3} \and Gennaro Raiola \inst{1,2}  \and Nikolaos Tsagarakis\inst{2} \and Marco Roveri\inst{3} \and  Enrico Mingo Hoffman\inst{4} \and Arash Ajoudani\inst{2}}


\authorrunning{Federico Rollo et al.} 

\institute{
Autonomous Systems, Leonardo Labs, Genoa, Italy, \email{\{name.surname\}.ext@leonardo.com}
\and
HHCM \& HRII, Istituto Italiano di Tecnologia, Genoa, Italy, \email{\{name.surname\}@iit.it}
\and
Industrial Innovation, DISI, Università di Trento, Trento, Italy, \email{\{name.surname\}@unitn.it}
\and
Université de Lorraine, CNRS, Inria, LORIA, Villers-lès-Nancy, France \\
\email{enrico.mingo-hoffman@inria.fr}
}

\maketitle              

\begin{abstract}
    Robots with a high level of autonomy are increasingly requested by smart industries. A way to reduce the workers' stress and effort is to optimize the working environment by taking advantage of autonomous collaborative robots.
A typical task for Human-Robot Collaboration (HRC) which improves the working setup in an industrial environment is the \textit{"bring me an object please"} where the user asks the collaborator to search for an object while he/she is focused on something else.
As often happens, science fiction is ahead of the times, indeed, in the \textit{Iron Man} movie, the robot \textit{Dum-E} helps its creator, \textit{Tony Stark}, to create its famous armours. The ability of the robot to comprehend the semantics of the environment and engage with it is valuable for the human execution of more intricate tasks. 
In this work, we reproduce this operation to enable a mobile robot with manipulation and grasping capabilities to leverage its geometric and semantic understanding of the environment for the execution of the \textit{Bring Me} action, thereby assisting a worker autonomously. Results are provided to validate the proposed workflow in a simulated environment populated with objects and people. This framework aims to take a step forward in assistive robotics autonomy for industries and domestic environments.  

\keywords{Semantic Loco-manipulation, Human-Robot Collaboration, Semantic Mapping, Assistive Robotics}

\end{abstract}

\section{Introduction} \label{sect:intro}
    Industries are increasingly demanding robotization to improve efficiency and simplify the complex tasks performed by human operators \textit{e.g.} by cooperation and collaboration. Numerous examples of Human-Robot Collaboration tasks can be found in the literature and industry \cite{vysocky2016human} \cite{lorenzini2023ergonomic}. For instance, in \cite{rollo2023followme}~\cite{rollo2023carpe}, the authors introduced a person-following application based on visual re-identification and gesture detection to assist humans.
Building on the work presented in \cite{rollo2023artifacts}, we can detect and localize objects in the environment and utilize this map to interact with these artifacts.\\
The primary contribution of this paper is the proposal of an easy-to-use framework for a mobile robotic collaborator to complete the \textit{"bring me"} service, leveraging semantic knowledge to accomplish the task. This work represents a step towards establishing a more comprehensive and adaptable application that can dynamically respond to the worker's needs to provide autonomous assistance.\\
The methodology is systematically presented in sect.~\ref{sect:method}, and in sect.~\ref{sect:experiments} the experiments and results are exposed and discussed. Finally, in sect.~\ref{sect:conclusion} the conclusive statements are reported along with some directions for future improvements.

\begin{figure}
    \centering
    \includegraphics[width=\linewidth]{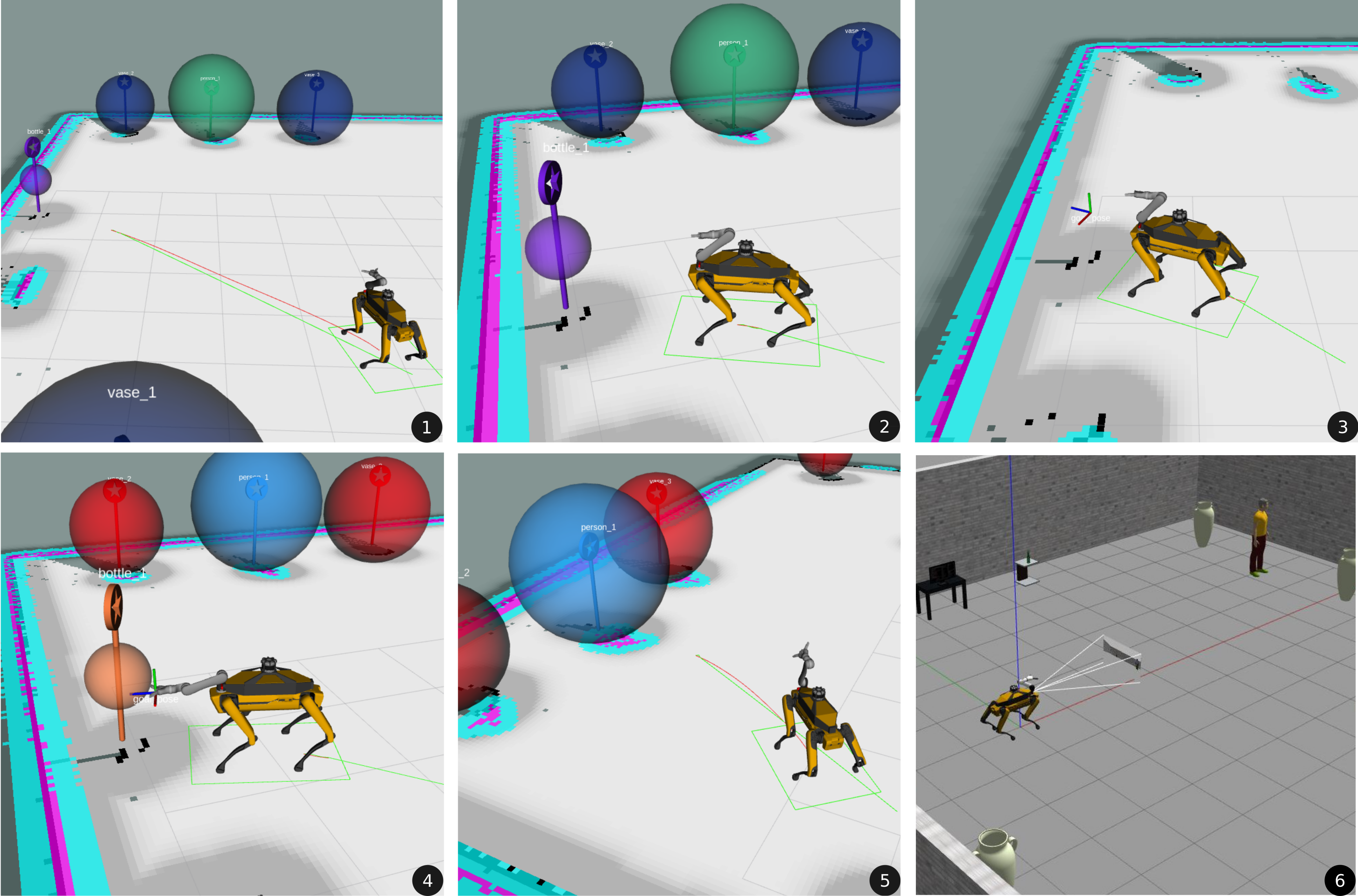}
    \caption{Autonomous actions held during the execution of the application. Image 1 represents the navigation towards the requested object. In image 2
    the object is approached, while in image 3 the object pose is estimated. In images 4 and 5 the object is picked and brought to the person. Image 6 shows the whole simulated world.}
    \label{fig:pipeline}
\end{figure}

\section{Method} \label{sect:method}
    To manage the application workflow we used a Behaviour Tree\footnote{Behaviour tree library: \url{https://py-trees.readthedocs.io/en/devel/}} (BT).
The BT is composed of a sequence of 6 actions: \textit{navigate\_to\_object}, \textit{approach\_object}, \textit{pose\_estimation}, \textit{pick\_object}, \textit{bring\_to\_user} and \textit{release\_object} and a safety action \textit{abort} which is activated when one of the other actions fails. Most of these actions are represented in fig.~\ref{fig:pipeline}. The BT awaits till the user makes a request, which is defined as the string associated with the artifact, \textit{e.g.}, bottle\_1. After the request is received the BT starts ticking and executes the behaviours hereafter explained.
\par
\textbf{navigate\_to\_object}:
the rough positions of both the requested object and the target person are already known, as documented in the previous mapping process~\cite{rollo2023artifacts}. This action involves taking the object's position and sending a goal to the robot in an obstacle-free region in front of the required object, allowing the robot to calculate its path and navigate towards the object.
\par
\textbf{approach\_object}:
the object approaching is necessary to facilitate a smooth picking action by refining the robot's position. This refinement is required due to the imprecise goal position received in the previous step. During this phase, the robot approaches the artifact, positioning it at the centre of the image and reducing the distance ensuring that the bottle falls within its manipulability space.
To accomplish this, the robot employs an instance segmentation network to segment the object and uses the resulting mask to crop both the RGB and Depth images. Thereafter, a point cloud is built using the masked depth image and the camera's proprioceptive parameters. For each non-zero pixel of the masked depth, the translation vector $\mathbf{^Bt_p}$ from the robot base frame $\mathbf{B}$ to the point $\mathbf{p}$ is computed as $\mathbf{^Bt_p} = \mathbf{^Bt_C} + \mathbf{^Ct_p}$, where $\mathbf{^Bt_C}$ is the translation vector from the base frame $\mathbf{B}$ to the camera frame $\mathbf{C}$ and $\mathbf{^Ct_p}$ is the translation vector from camera frame $\mathbf{C}$ to the generic 3D point $\mathbf{p}$ and is computed as:
\begin{equation} \label{eq:camera_pcd}
    \mathbf{^Ct_p}
    =    
    \begin{bmatrix}
        \frac{1}{f_x} & 0  & -\frac{p_x}{f_x} \\
        0  & \frac{1}{f_y} & -\frac{p_y}{f_y} \\
        0  & 0  & 1 
    \end{bmatrix}
    \begin{bmatrix}
        u \\ v \\ 1 
    \end{bmatrix}
    d_C\text{ ,}
\end{equation}
 where $f_x$, $f_y$, $p_x$ and $p_y$ are the camera intrinsic parameters \textit{i.e.}, the focal lengths and principal points of the camera along $x$ and $y$ image axis, $u$ and $v$ are the depth pixel position along $x$ and $y$ axis and $d_C$ is the depth value of the considered pixel.
With this point cloud, we can compute the object's centroid $\mathbf{o}$ expressed in the base frame $\mathbf{B}$, denoted as $\mathbf{^Bt_o}$, and two key metrics: the 3D Euclidean distance to the robot base, denoted as $d_{obj}$, and the heading angle between the robot and the object, referred to as $\theta_{obj}$, with the equation:
\begin{equation} \label{eq:h_angle}
    \theta_{obj} = atan2(y, x)
\end{equation}
where $x$ and $y$ are the coordinates of the centroid $\mathbf{^Bt_o}$ (the $xy$ plane is parallel to the ground floor), and $atan2$ is the arc tangent function which takes into account the quadrant of the tangent.
We then used two linear proportional controllers (\textit{e.g.}, eq.~\ref{eq:controller}) to regulate the robot base position sending 
velocity commands (linear velocity along the x-axis $v_x$, angular velocity around the z-axis $\omega_z$) using as reference the distance $d_{opt}$ and the heading angle $\theta_{opt}$.
\begin{equation} \label{eq:controller}
    \xi = K (\chi - \chi_{opt})\text{ ,}
\end{equation}
where $\xi$ is the linear $v_x$ or angular $\omega_z$ velocity output, $K$ is the proportional gain, $\chi$ is the current distance $d_{obj}$ or the heading angle $\theta_{obj}$ and $\chi_{opt}$ is the optimal distance $d_{opt}$ or angle reference $\theta_{opt}$.

\textbf{pose\_estimation}:
once the robot is well positioned, the homogeneous transformation $\mathbf{^BH_{G}}$ of the grasping pose $\mathbf{G}$ expressed in the base frame $\mathbf{B}$, is computed. Using the instance segmentation neural network and the RGB-D camera we compute the translation vector $\mathbf{^Bt_G}$ from base frame $\mathbf{B}$ to the grasping pose $\mathbf{G}$ in the same way as $\mathbf{^Bt_o}$ in the \textit{approach\_object} behaviour. We then set the rotation matrix $\mathbf{^BR_{G}}$ between the robot base $\mathbf{B}$ and the grasping pose $\mathbf{G}$ equal to an identity matrix $\mathbf{I_{3x3}}$ to obtain the homogeneous transformation:
\begin{equation}
    \mathbf{^BH_{G}} = 
    \begin{bmatrix}
        \mathbf{^BR_{G}} & \mathbf{^Bt_G} \\
        0\ 0\ 0 & 1
    \end{bmatrix}
\end{equation}

\textbf{pick\_object}: 
in this phase, to move the robot gripper in the final pose, the robot receives the desired grasping homogeneous transformation $\mathbf{^BH_{G}}$ and roto-translates it in the end-effector frame $\mathbf{E}$ as:
\begin{equation}
    \mathbf{^EH_{G}} = \mathbf{^EH_{B}}  \mathbf{^BH_{G}}
\end{equation}

In our case, the simulation uses an optimization-based whole-body inverse dynamics controller~\cite{raiola2022wolf} to move the arm towards the target pose. Still, this step depends on the specific robot configuration used. 

When the robotic end-effector is in the picking position, the gripper can grasp to pick up the artifact, and the arm can then move to a folded position for transportation.

\textbf{bring\_to\_user}:
once the artifact is grasped, the robot brings it to the user navigating in the environment.

\textbf{release\_object}:
finally, when the robot reaches the user position it gives the requested object to the user. 

\textbf{abort}:
the abort status is essential for managing unexpected behaviours. All goals or requests sent are halted in this state, and the robot enters an idle state, awaiting further commands. The mission is deleted, and the robot promptly reports the error to the user.

\section{Experiments} \label{sect:experiments}
Experiments were conducted in simulation using WoLF~\cite{raiola2022wolf}, a framework that allows the simulation and control of a quadruped robot with an arm attached to it. The experiments were run on a notebook with an Intel® Core™ i9-11950H processor and an NVIDIA Geforce RTX 3080 Laptop GPU. For the robot motion planning and navigation in the environment, we used the ROS Navigation Stack\footnote{ROS Navigation Stack: \url{http://wiki.ros.org/navigation}} and YOLOv8\footnote{YOLOv8: \url{https://github.com/ultralytics/ultralytics}} with pre-trained weights as the instance segmentation network. The code is implemented in \textit{C++} and \textit{Python} and integrated with \textit{ROS}.

In the experiments, the robot is asked to bring an object to one of the people in the simulation (e.g. bring a bottle to the person with id 1). The experiment is repeated for different world configurations where the people and the objects are randomly moved (see an example in image 6 of fig.~\ref{fig:pipeline}). The proposed pipeline always completed the task following the steps presented in sect.~\ref{sect:method}. A video of an experiment is provided at \url{https://youtu.be/6Qk1eHCrCqo}.
    
\section{Conclusion} \label{sect:conclusion}
    In this article, we have introduced a semantic loco-manipulation framework for object retrieval. This application uses high-level semantic scene understanding to enable a robot assistant to search for and bring objects that are not nearby and whose locations may be unknown to the user. It is a foundational platform for enhancing robotic assistance within industrial environments, where robots work alongside human operators and actively participate in the workplace community.
Several future developments for this framework include the generalization of the pose estimation to handle more complex situations and objects, making experiments on a real robot and improving the reactivity of the BT, \textit{e.g.} if the robot loses the grasp of the object the BT restart autonomously from the \textit{pose\_estimation} behaviour.

\bibliography{biblio}
\bibliographystyle{plain}
\end{document}